\documentclass[letterpaper]{article} 
\usepackage{aaai24}  
\usepackage{times}  
\usepackage{helvet}  
\usepackage{courier}  
\usepackage[hyphens]{url}  
\usepackage{graphicx} 
\usepackage{natbib}  
\usepackage{caption} 
\usepackage{algorithm}
\usepackage{algorithmic}
\usepackage{makecell}
\usepackage{multirow}
\usepackage{graphicx}
\usepackage{subcaption}
\usepackage{booktabs}
\usepackage{amsmath}
\usepackage{amssymb}

\usepackage{newfloat}
\usepackage{listings}
\DeclareCaptionStyle{ruled}{labelfont=normalfont,labelsep=colon,strut=off} 
\floatstyle{ruled}
\newfloat{listing}{tb}{lst}{}
\floatname{listing}{Listing}
\usepackage{color, colortbl} 
\definecolor{Gray}{gray}{0.9}
\definecolor{LightCyan}{rgb}{0.88,1,1}

\title{UMIE: Unified Multimodal Information Extraction with Instruction Tuning}
\author{
Lin Sun\textsuperscript{\rm 1}\thanks{These three authors contributed equally to this work.},
Kai Zhang\textsuperscript{\rm 2}$^*$,
Qingyuan Li\textsuperscript{\rm 3}$^*$,
Renze Lou\textsuperscript{\rm 4}
}
\affiliations{
     \textsuperscript{\rm 1}Department of Computer Science, Hangzhou City University, China\\
     \textsuperscript{\rm 2}Department of Computer Science and Engineering, Ohio State University, USA\\
    \textsuperscript{\rm 3}College of Computer Science and Technology, Zhejiang University, China\\
    \textsuperscript{\rm 4}Department of Computer Science and Engineering, Pennsylvania State University, USA\\
   sunl@hzcu.edu.cn, liqingyuan@zju.edu.cn
    
}

\usepackage{bibentry}

\begin{document}

\maketitle

\begin{abstract}

Multimodal information extraction (MIE) gains significant attention as the popularity of multimedia content increases.
However, current MIE methods often resort to using task-specific model structures, which results in limited generalizability across tasks and underutilizes shared knowledge across MIE tasks.
To address these issues, we propose UMIE, a unified multimodal information extractor to unify three MIE tasks as a generation problem using instruction tuning, being able to effectively extract both textual and visual mentions.
Extensive experiments show that our single UMIE outperforms various state-of-the-art (SoTA) methods across six MIE datasets on three tasks.
Furthermore, in-depth analysis demonstrates UMIE's strong generalization in the zero-shot setting, robustness to instruction variants, and interpretability.
Our research serves as an initial step towards a unified MIE model and initiates the exploration into both instruction tuning and large language models within the MIE domain.
Our code, data, and model are available at https://github.com/ZUCC-AI/UMIE.
\end{abstract}

\section{Introduction}
    
Information extraction (IE), a task aiming to derive structured information from unstructured texts, plays a crucial role in the domain of natural language processing.
As the popularity of multimedia content continues to increase~\cite{Zhu2022MultiModalKGSurvey}, multimodal information extraction (MIE) has drawn significant attention from the research community~\cite{zhang2018adaptive, Chen2022HVPNeT, Li2022CLIP-Event}.
MIE aims to deliver structured information of interest from multiple media sources such as textual, visual, and potentially more.
It is considered a challenging task due to the inherent complexity of media formats and the necessity to bridge cross-modal gaps, where traditional text-based IE methods often struggle.

MIE includes multimodal named entity recognition (MNER)~\cite{moon2018multimodal, zhang2018adaptive, sun2021rpbert}, multimodal relation extraction (MRE)~\cite{zheng2021mnre, Wang2022MoRE}, and multimodal event extraction (MEE)~\cite{li2020WASEandM2E2, Li2022CLIP-Event}.
Current methods for usually focus on a specific task mentioned above, which mostly uses a task-specific model structure with dataset-specific tuning for the task at hand.
Such a paradigm leads to a few limitations:
Firstly, it results in a lack of generalizability as they often overfit the patterns of a specific task or even dataset.
Secondly, the need to design, train, and maintain a separate model for each task is time-consuming, impeding the progress of deploying practical multimodal systems at scale.
Lastly, due to their independent training nature, these methods fail to effectively leverage the shared knowledge across different MIE tasks, undermining the performance of each task.

\begin{figure}[t]
	\small
		\centering
			\includegraphics[width=0.4\textwidth]{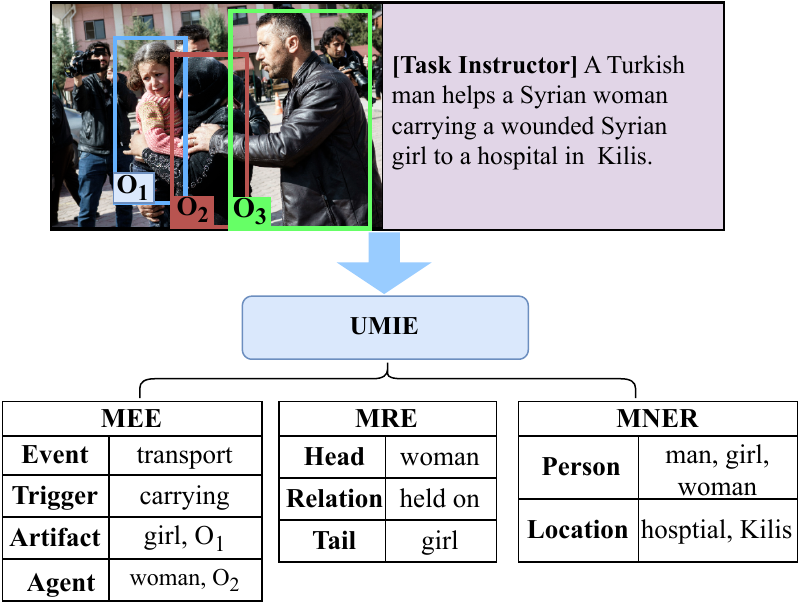}
		\caption{ Unifying three key MIE tasks in a single multimodal model.
  Given a task instructor, UMIE performs the corresponding task by extracting textual and visual mentions (MNER and MEE) or inferring the relationship between two given mentions (MRE). $O_1$, $O_2$, and $O_3$ are visual objects.}
		\label{fig:motivation}
\end{figure}

To address these challenges, in this work, we propose a unified multimodal information extractor (UMIE), a single model that unifies different MIE tasks as generation problems with instruction tuning~\cite{Ouyang2022InstructGPT, Wang2022SuperInstructions}.
As shown in Figure~\ref{fig:motivation}, given the same text and image, UMIE can perform various MIE tasks following different task instructors and generate corresponding structured outputs.
In particular, UMIE enables both extraction of textual span and visual objects (e.g., MEE), which is rarely considered by previous MNER and MRE works.

Specifically, we repurpose all MIE datasets and train a UMIE model to perform each task in a generation fashion by following corresponding instructions.
In addition, we design a visual encoder and a gated attention module to dynamically integrate visual clues from images for robust cross-modal feature extraction.
Consequently, our singular UMIE model outperforms various SoTA methods on each of the six MIE datasets across all three tasks, demonstrating the effectiveness of our framework.
Furthermore, in-depth analysis showcases strong generalization ability and robustness to instructions.
Our contributions are summarized as follows:
\begin{itemize}
    \item We present UMIE, an end-to-end model that unifies MIE tasks into a generation problem.
    To the best of our knowledge, this study is the first step towards unified MIE and initiates exploration into both instruction tuning and large language models in the MIE field.
    \item We propose a gate attention module to dynamically utilize visual features and conduct in-depth quantitative analysis on each MIE task, shading light into the role of visual features and validating our gate mechanism's effectiveness and importance in our model.
    \item Extensive experiments show the effectiveness across six datasets of three MIE tasks, generalization ability in the zero-shot setting, and instruction robustness of our UMIE model.
    We will release all MIE datasets with standard format and models trained on them, as a benchmark and starting point for future studies in this area of unified multimodal information extraction.
    
\end{itemize}

\section{Related Work}

\smallskip
\noindent
\textbf{Multimodal Named Entity Recognition (MNER).}
The MNER task aims to recognize mentions and classify them into predefined categories from texts, using additional visual clues provided in images. 
Pioneering work~\cite{moon2018multimodal, lu2018visual, zhang2018adaptive} focuses on utilizing visual features for improved representation learning.
Several works~\cite{arshad2019aiding, yu2020improving} point out that failure recognitions are due to unrelated images that corrupt visual attention and mislead entity recognition. 
As a remedy, Yu et al.~\shortcite{yu2020improving} propose the Unified Multimodal Transformer (UMT), in which a visual gate dynamically utilizes visual information for final representation.
Additionally, RIVA~\cite{sun-etal-2020-riva} and RpBERT~\cite{sun2021rpbert} explicitly predict the relevance of the given image and text, using text-image relationship binary classification as an auxiliary task, thus addressing the misleading issue introduced by irrelevant visual content.

Recently, ITA~\cite{wang2022ita} capitalizes on the objects in the given image and inputs with all text, thus leading to a unified representation and a better attention mechanism over texts. 
MNER-QG~\cite{Jia2023MNER-QG} frames MNER as a machine reading comprehension task, querying the models for entity recognitions. MoRe~\cite{Wang2022MoRE} enhances the MNER model with related texts obtained via information retrieval techniques~\cite{Zhang2023LED, Shen2023UnifieR}.

\smallskip
\noindent
\textbf{Multimodal Relation Extraction (MRE).}
MRE~\cite{zheng2021mnre, zheng2021multimodal} aims to identify the semantic relationships between two entities based on the given text image pair.
HVPNeT~\cite{Chen2022HVPNeT} fuses image information as a prefix for better text representation.
MoRe~\cite{Wang2022MoRE} retrieves related texts from the entire Wikipedia dumps for boosting both MNER and MRE performance.
Follow-up work~\cite{Hu2023CrossRetrieval} further retrieves relevant images to the object, text, and image for better retrieval augmentation~\cite{Xie2023LLM-KnowledgeConflict, Yue2023AttrScore}.
However, they involve time-consuming retrieval over a large-scale collection and require an external knowledge base.
Therefore, these two retrieval-augmented methods cannot be directly compared with our work, which focuses on model development itself.

\smallskip
\noindent
\textbf{Multimodal Event Extraction (MEE).} MEE aims to extract events (i.e., \textbf{Event Detection}) and arguments for the event (i.e., \textbf{Event Argument Extraction}) from multiple modalities.
VAD~\cite{Zhang2017VAD} leverages additional images to enhance event extraction by alleviating ambiguity in the text modality.
Tong et al.~\shortcite{Tong2020DRMM} propose DRMM, which recurrently uses related images from a constructed supplementary image set for augmenting text-only event detection, thereby improving the disambiguation of trigger words. 
Li et al.~\shortcite{li2020WASEandM2E2} propose M$^2$E$^2$ data and obtain weak supervision from text-only, image-only, and image-caption data to encode visual and textual data into a joint representation space for extraction. 
Follow-up work~\cite{Li2022CLIP-Event} pre-trains a vision and language model in an event-level alignment with a contrastive learning fashion over a large event-rich dataset and generalize to the M$^2$E$^2$ dataset.
Recently, Unicl~\cite{liu2022multimedia} proposes a unified contrastive learning framework to bridge the modality gaps. 

Notably, MRE and MNER focus on leveraging visual information to enhance extraction from texts, rather than extraction over images while MEE may extract visual objects as arguments in an event.
To the best of our knowledge, our proposed UMIE unifies the extraction from text and image in the multimodal IE domain for the first time.

\smallskip
\noindent
\textbf{Unified Information Extraction.} 
Despite the diversity and heterogeneity of information extraction (IE) tasks, several works unify these IE tasks as text-to-structure generation~\cite{Lu2022UIE} or semantic matching~\cite{Lou2023USM}.
However, these models focus on text-only IE tasks.
Our UMIE model solves multimodal IE tasks universally.

Meanwhile, instruction tuning~\cite{Lou2023Instruction} fine-tunes models to follow instructions, showing unprecedented zero-shot generalization abilities when models perform unseen tasks given new instructions~\cite{Sanh2022T0, Ouyang2022InstructGPT, ChungFlanT5}.
However, as pointed out by recent works~\cite{Zhang2023QA4RE}, such instruction-tuned LLMs fail to achieve decent results in IE tasks due to the low incidence of these tasks during instruction tuning.
Additionally, Chen et al.~\cite{Chen2023CoT-MNER} provide evidence that LLMs like ChatGPT~\footnote{https://openai.com/blog/chatgpt} and GPT-4~\cite{openai2023gpt4} yield poor results on both MNER and MRE tasks.
Given high long-term costs and potential risks of using test data for training, such LLMs are not ideal choices for information extraction tasks.
Therefore, open-sourced IE-specific models are more transparent, effective, and cost-efficient in practice, which is the focus of our work.

\section{Unified Multimodal Information Extractor}

\subsection{Model Overview}
\begin{figure}[t]
    \centering
        \includegraphics[width=0.45\textwidth]{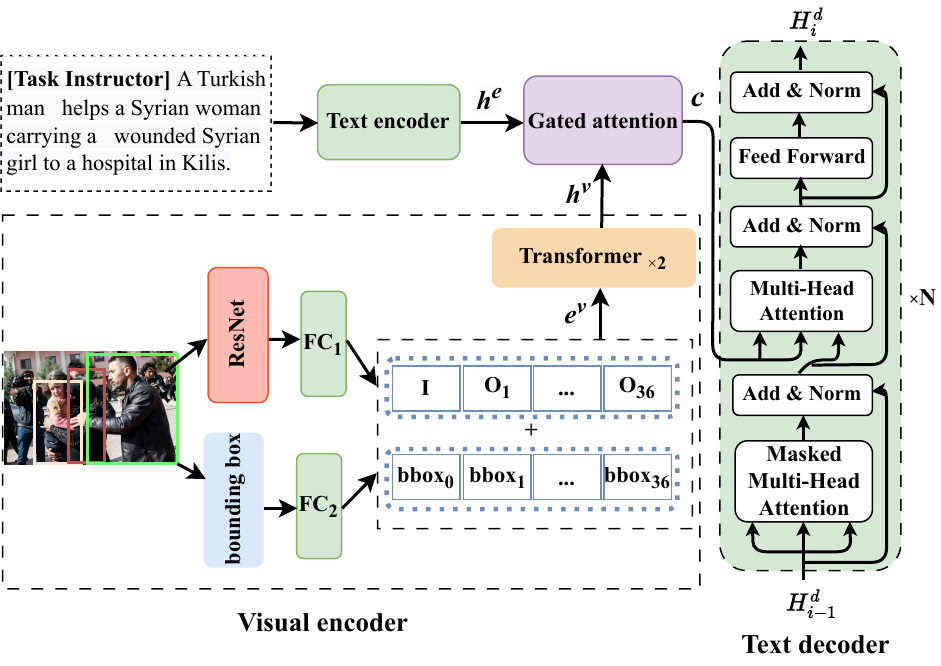}
    \caption{Illustration of the UMIE model.
    The visual encoder encodes an image and objects into features that are dynamically integrated with textual features in the gated attention module and the text decoder generates information extraction results autoregressively.}
    \label{fig:model}
\end{figure}

 As shown in Figure~\ref{fig:model}, the Unified Multimodal Information Extractor (UMIE) consists of four major modules:
 1) Text encoder for instruction-following and text comprehension;
 2) Visual encoder for visual representation;
 3) Gated attention for cross-modal representation;
 4) Text decoder for information extraction. 
The UMIE model utilizes a transformer-based encoder-decoder architecture to perform MIE and generate structured outputs in an auto-regressive fashion.
For the textual input prefixed with a task instructor, we use a text encoder to generate text representations.
For the image, we equip it with visual understanding abilities via our proposed visual encoder and gated attention mechanism for dynamic visual clue integration, the details of which will be introduced in the following sections.
Finally, we employ a text decoder to generate the structural results for MIE tasks.
In this work, we utilize FLAN-T5~\cite{ChungFlanT5}, a Transformer-based encoder-decoder large language model (LLM), to initialize the structure and parameters of the encoder and decoder components of the UMIE model.


Specifically, the UMIE computes the hidden vector representation $h^e = \{h_1, \dots,h_n\} \in \mathbb R^{n \times d_t}$ of the input text $\{w_1, \dots,w_n\}$ where $d_t$ is the dimension of token embeddings:
\begin{equation}
\label{eqn: text_feat}
h^e= \text{Text-Encoder}(w_1,\dots,w_n).
\end{equation}


In the gated attention module, we use the textual feature $h^e$ as the query and the visual feature $h^v$ as the key and value for the cross-attention computation.
For dynamic incorporation of the text-aware visual representation, we design a gated signal $g$ to control the final output of the cross-modal representation $c$ (refer to Section ``Gated Attention Module''):
\begin{equation}
\label{eqn: cross_modal_feat}
    c = \text{Gated-Attention}(h^e,h^v).
\end{equation}

Using the cross-modal representation $c$, the text decoder generates the output structure in an autoregressive manner, starting with the input of the start token $<$$s$$>$ as the initial step.
Such a generation process ends with the end token $<$$/s$$>$.
At step $i$, the text decoder represents the state $H_i^d$ conditioned on the cross-modal state $c$ and previous states $[H_1^d, \dots, H_{i-1}^d]$.
Formally,
\begin{equation}
    H_i^d = \text{Text-Decoder}([c~; H_1^d, \dots, H_{i-1}^d]),
\end{equation}
The text decoder consists of $N$-layer Transformers, additionally inserting a third sub-layer that performs multi-head attention over the output $c$ of the gated attention module, a similar approach as described in~\cite{vaswani2017attention}.
Based on the states $[H_1^d, \dots, H_{i}^d]$, we can decode a text sequence via a linear projection and softmax function.

\subsection{Visual Encoder}
In MIE tasks, the associated image typically offers a valuable visual clue, directing the models towards the information of interest. 
Therefore, to incorporate semantic knowledge, we encode the local objects.
Furthermore, we consider the image's global feature for additional context information.
By collecting both global and regional images, our visual encoder has the potential to extract more visual clues, leading to improved information utilization and more accurate extraction.

Specifically,  we utilize semantic objects detected by off-the-shelf visual grounding toolkit~\cite{tan2019lxmert}.
This toolkit provides up to 36 local region-of-interest (RoI) features as well as bounding box (bbox) coordinates which are $x$ and $y$ coordinates of the top left corner and the bottom right corner of the object rectangle.
If the number of objects are less than 36, the remaining object features are padded with zeros.
To unify the RoI inputs, we rescale the image and a series of visual objects to a size of $224 \times 224$ pixels.

Given an image $I$ and its objects $\{O_1, \dots, O_{36}\}$, the backbone ResNet-101~\cite{He2016ResNet} extracts visual features as $f = \{f_I, f_{O_1}, \dots, f_{O_{36}}\} \in \mathbb R^{37 \times d_v}$.
Then two fully connected (FC) layers are applied to visual features $f_i$ and their corresponding bounding box coordinates $bbox_i$ to obtain visual embeddings $e^v_i$:
\begin{equation}
\label{eqn: visual_embedding}
e^v_i = \text{FC}_1(f_i) + \text{FC}_2(bbox_i),
\end{equation}
where $\text{FC}_1\in\mathbb R^{d_v \times d_t}$ and $\text{FC}_2\in\mathbb R^{4 \times d_t}$.
The visual embeddings $e^v$ are further processed by a 2-layer Transformer~\cite{vaswani2017attention} to obtain the final visual features $h^v \in \mathbb R^{37 \times d_t}$:
\begin{equation}
\label{eqn: visual_feat}
h^v = \text{Transformer}_{\times 2}(e^v).
\end{equation}
The features $h^v$ can represent the integration of global image and RoI local visual information.


\begin{table*}[t]
\centering
\small
\begin{tabular}{ccl} 
\toprule 
 \multicolumn{2}{c}{\textbf{Task}}& \multicolumn{1}{c}{\textbf{Task Instructor}} \\
 \midrule
 \multicolumn{2}{c}{MNER}&  Please extract the following entity type: person, location, miscellaneous, organization.\\
\midrule
 \multicolumn{2}{c}{MRE}& \makecell[l]{Please extract the following relation between [head] and [tail]: part of, contain, present in, none, held on, \\
member of, peer, lace of residence, locate at, alternate names, neighbor, subsidiary, awarded, couple
\\parent,  nationality, place of birth, charges, siblings, religion, race.}\\
\midrule
 \multirow{2}{*}{MEE}&MED& \makecell[l] {Please extract the following event type:\\ arrest jail, meet, attack, transport, demonstrate, phone write, die, transfer.}\\\cmidrule{2-3}
 &MEAE &\makecell[l] {Given an [TYPE] event, please extract the corresponding argument type: [Argument Roles]. \\(e.g., \textit{given an arrest jail event, please extract the corresponding argument type: person, place, agent...})}\\
\bottomrule
\end{tabular}
\caption{The description of task instructors. As shown in the last row of the table, the specific instruction of MEAE will be determined by the event type detected by MED.}
 \label{tab:task-instructor}
\end{table*}

\subsection{Gated Attention Module}
\begin{figure}[t]
    \small
    \centering
    \includegraphics[width=0.2\textwidth]{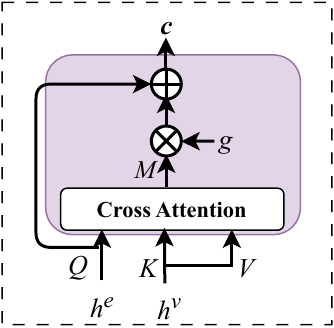}
    \caption{Illustration of the gated attention module.}
		\label{fig:GA}
\end{figure}

Previous work illustrated the potential irrelevance of images~\cite{sun2021rpbert} or inductive bias~\cite{yu2020improving} introduced by the images.
To alleviate such an issue, we design a gate attention module shown in Figure~\ref{fig:GA} to dynamically integrate visual and textual features and the gate module controls the contribution of visual features for the text decoder.

\smallskip
\noindent
\textbf{Visual-textual Cross Attention.}
In the attention module, the visual features $h^v$ serve as the key and value, while the textual representations $h^e$ from the text decoder serve as the query.
In this way, the textual representations are used to attend to the most relevant visual features, generating the text-aware visual representation $M \in \mathbb R^{n \times d_t}$:

\begin{eqnarray}
    \label{eqn: multimodal-feat}
    M&=&\text{Cross-Attention}(Q=h^e,K=h^v,V=h^v) \nonumber \\
		~&=&\text{softmax}(\frac{Q K^T}{\sqrt{d_t}})V,
\end{eqnarray}
where $Q$, $K$, and $V$ respectively represent query, key, and value.
The softmax function is applied on the matrix multiplication of $Q$ and the transpose of $K$, divided by the square root of the model dimension and used as weights multiplied with $V$.

\smallskip
\noindent
\textbf{Gate Control.}
To control the text-aware visual feature $M$, we design a gate signal $g$ to denote the contribution of the visual feature in the text decoding process.
The gate signal $g$ is dynamically calculated by the following equation:
\begin{equation}
    \label{eqn: gate}
		g = \text{LeakyReLU}(\overline{K}~\overline{Q}^T),
\end{equation}
where $\overline{K}$ is the mean of vectors in $K$ and $\overline{Q}$ is the mean of vectors in $Q$.

Finally, the gate signal $g$ controls the weight of text-aware visual feature $M$ to deliver the cross-modal information $c \in \mathbb R^{n \times d_t}$ by textual features $h^e$ and gated residual visual features $g \cdot M$, 
\begin{eqnarray}
     \label{eqn: cross-feature}
      c&=&\text{Gated-Attention}(h^e,h^v) \nonumber \\
			~&=&h^e + g \cdot M.
\end{eqnarray}
Such cross-modal representation is used for structure generation by the text decoder in an auto-regressive fashion, thus dynamically incorporating visual features for multimodal information extraction.

\begin{table*}[!ht]
\centering

\resizebox{0.96\linewidth}{!}{
\begin{tabular}{cccc}
    \toprule 
    \textbf{Task} & \textbf{Image} & \textbf{Text} & \textbf{Output}\\
    \midrule
    MNER 
    & \makecell[c]{\begin{minipage}[b]{0.3\columnwidth}
    \centering
    {\includegraphics[width=0.9\textwidth]{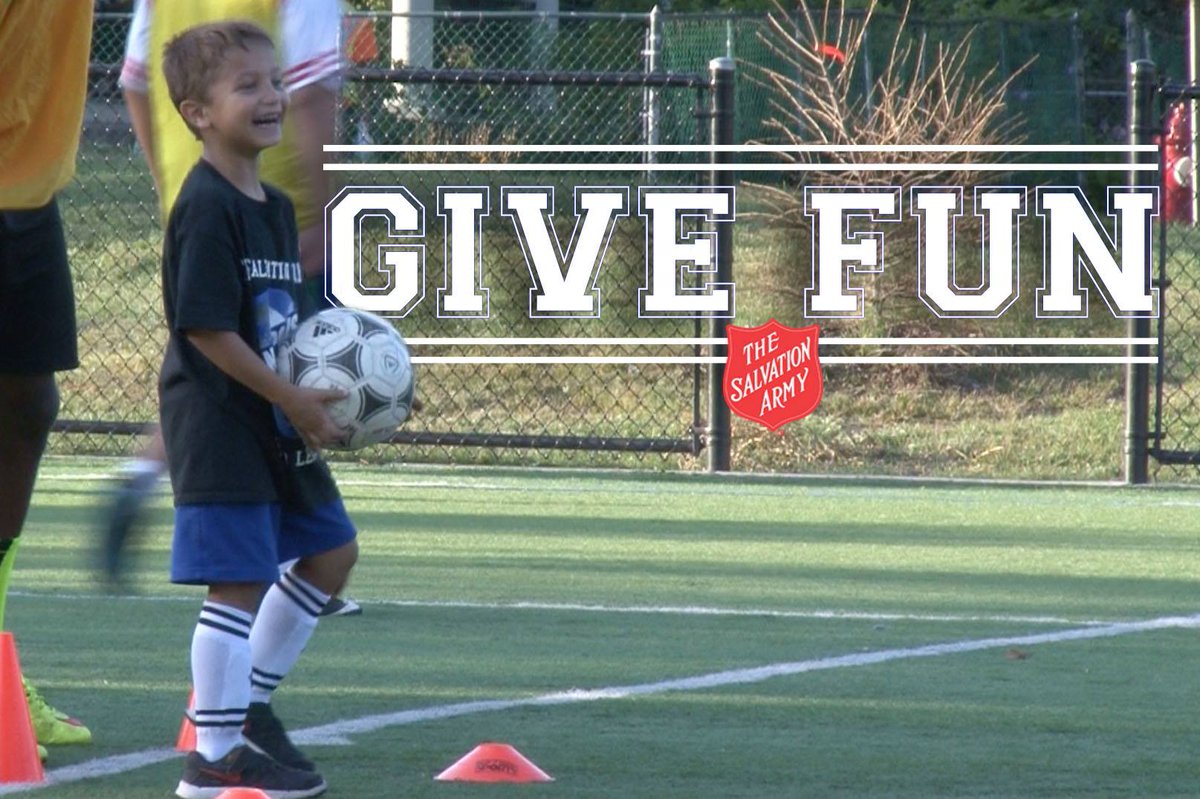}}
    \end{minipage}}
    & \makecell[l]{\textbf{[MNER Task Instructor]} Support the \\Salvation Arm free kids soccer program \\by donating a ball or two this summer ! \\
     UEFAcom \# uclfinal} & \makecell[l]{\textit{\textbf{Person}}, kids  $<$$\text{spot}$$>$ \textit{\textbf{Organization}}, Salvation Army } \\
    \midrule
    MRE 
    & \makecell[c]{\begin{minipage}[b]{0.3\columnwidth}
    \centering
    {\includegraphics[width=0.9\textwidth]{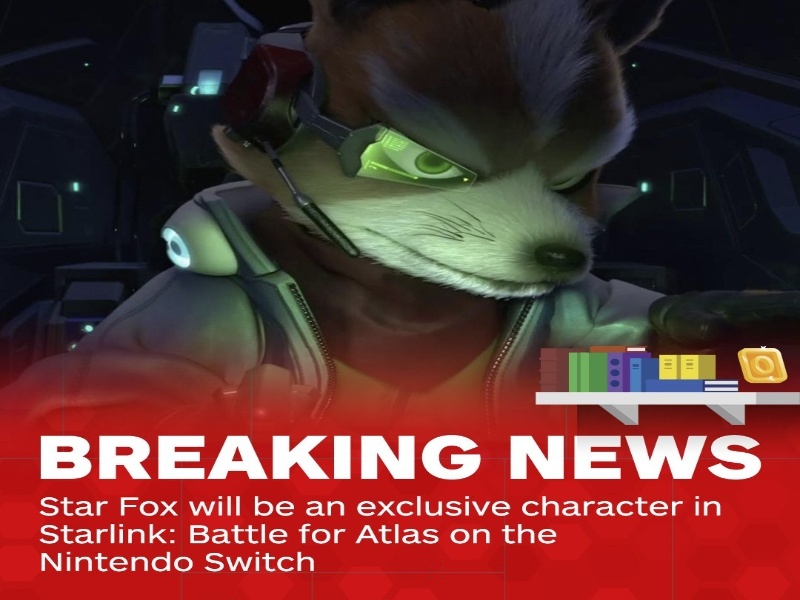}}
    \end{minipage}}
    &  \makecell[l]{\textbf{[MRE Task Instructor]}  Star Fox will be  \\ an exclusive character in Starlink. } &  \makecell[l]{Star Fox, \textit{\textbf{member of}}, Starlink}\\
    \midrule

    \multirow{4}{*}{MEE} & \multirow{2}{*}{\makecell[c]{\begin{minipage}[b]{0.3\columnwidth}
    \centering
    {\includegraphics[width=1.0\textwidth]{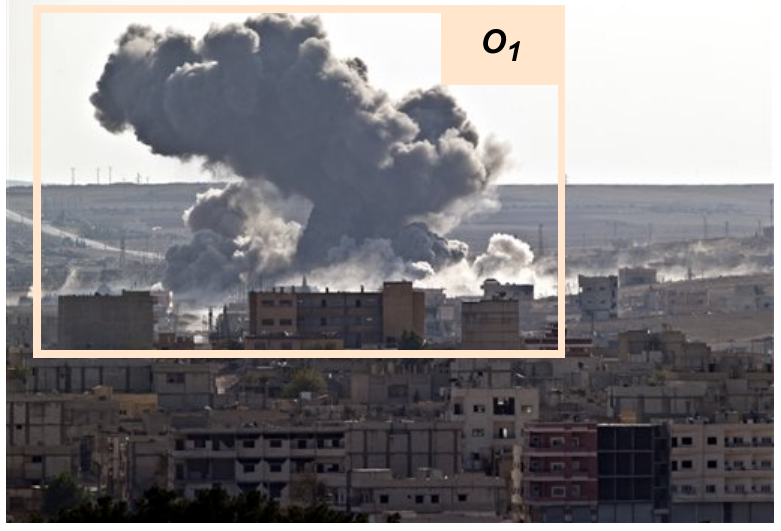}}
    \end{minipage}}}
   & \makecell[l]{\textbf{[MED Task Instructor]} Smoke rises over \\the Syrian city of Kobani , following a US \\led coalition airstrike, seen from outside Suruc.}
    & \makecell[l]{\textit{\textbf{Attack}}, airstrike} \\\cmidrule{3-4}
   & & \makecell[l]{\textbf{[MEAE Task Instructor]} Smoke rises over \\the Syrian city of Kobani , following a US \\led coalition airstrike, seen from outside Suruc.} & \makecell[l]{\textit{\textbf{Attacker}}, coalition $<$$\text{spot}$$>$ \textit{\textbf{Target}}, O$_1$}\\

    \bottomrule

    \end{tabular}

}
    \caption{The input and output format of three tasks. In particular, the MEE task is be conducted as two cascaded tasks, namely MED and MEAE, with the same input example.}\label{tab:in-out}  
\end{table*}

\subsection{Task Instructor and Text Decoding}
With multimodal understanding and generation abilities, UMIE could perform various multimodal information extraction tasks, given different instructions.
Tables~\ref{tab:task-instructor} and~\ref{tab:in-out} show our task instructors and input and output formats used for MNER, MRE, and MEE, where MEE is conducted as multimodal event detection (MED) and multimodal event argument extraction (MEAE) in a cascaded fashion.
Such instructors will be used as prefixes during both training and evaluation.
For MRE, UMIE simply decodes the triples, including the relationship.
For MNER, MED, and MEAE, we introduce $<$$\text{spot}$$>$ to separate the classes.
Specifically for event extraction tasks, we detect events by finding and identifying the triggers first by MED, based on which we further extract corresponding arguments by MEAE.



\section{Experiments}
We train the UMIE with instruction tuning on various MIE datasets and evaluate the model in both supervised learning and zero-shot settings.
In addition, we evaluate the robustness of instruction following of UMIE and showcase the unified extraction abilities of our model.

\subsection{Experiment Setup}

\begin{table}[t]
\small
\centering

    \begin{tabular}{l|l|lcc} 
        \toprule
        \textbf{Task}  & \textbf{Dataset} & \textbf{Train} &  \textbf{Dev} & \textbf{Test}\\  
        \midrule
        \multirow{3}{*}{MNER}  & Twitter-15 & 4,000 &  1,000 & 3,257\\
         & Twitter-17 & 2,848 &  723 & 723\\
         & SNAP &  3,971 &  1,432 & 1,459\\
         \midrule
        \multirow{2}{*}{MRE}&  MNRE-V1 & 7,824 & 975 & 1,282 \\
        & MNRE-V2 & 12,247 & 923 & 832 \\
        \midrule
        MEE & M$^2$E$^2$ & - &  - & 309\\
        \bottomrule
    \end{tabular}
        \caption{The statistics of six MIE datasets.}    \label{tab:dataset}
        \vspace{-1.5em}
\end{table}

\smallskip
\noindent
\textbf{Datasets.}
We train and evaluate UMIE on several datasets commonly used in MNER, MRE, and MEE tasks:
1) For MNER, we consider Twitter-15~\cite{zhang2018adaptive}, SNAP~\cite{lu2018visual}, and Twitter-17~\cite{yu2020improving} (a refined version of SNAP), all curated from the social media platform;
2) For MRE, we adopt the MNRE dataset~\cite{zheng2021mnre} constructed from the social media domain via crowdsourcing;
3) For MEE, following previous work~\cite{Tong2020DRMM}, we employ datasets such as ACE2005~\cite{walker2006ace}, SWiG~\cite{pratt2020grounded} for training, and the M$^2$E$^2$dataset for evaluation.
ACE2005 features textual event annotations with 33 event types and 36 argument types.
We utilize the M$^2$E$^2$ dataset from~\cite{li2020WASEandM2E2}, containing 245 documents labeled with parallel textual and visual events.
The M$^2$E$^2$ event schema aligns with 8 ACE types and 98 SWiG types.
The event instances are divided into 1,105 text-only events, 188 image-only events, and 385 multimedia events.
Due to partial overlap between the Twitter-17 and SNAP datasets, we have removed 319 Twitter-17 testing samples from the SNAP training set, while retaining 2848 Twitter-17 training samples, ensuring that no data overlaps between the training and test sets.
Detailed statistics of data are listed in Table~\ref{tab:dataset}.
We have reformatted all datasets into a standardized JSON format for training our UMIE model.

\smallskip
\noindent
\textbf{Training Configuration.}
We train our model by employing label smoothing and AdamW, with a learning rate of 5e-5 for FLAN-T5-large and 1e-4 for FLAN-T5-base.
The number of training epochs is set to 40.
All experiments are conducted on 8 NVIDIA A100 GPUs, each possessing a memory capacity of 40GB.
Due to GPU memory limitations, we use different batch sizes: 8 for FLAN-T5-large and 16 for FLAN-T5-base.
During the training process, we restrict the output of the text input to a maximum length of 256 and the generated length to 128.
We add the task instructor as the prefix in the default setup for FLAN-T5.

\smallskip
\noindent
\textbf{Evaluation Metrics.}
Among these four MIE tasks (MNER, MRE, MED, and MEAE), we conduct comprehensive comparisons between our UMIE and state-of-the-art multimodal approaches in each task.
We use the reported performances from the respective papers for comparison.
In terms of the evaluation metrics, we follow the F1 metric commonly used in each task as described in the respective papers.

 \begin{table*}[t]
    \centering
    \small
    \begin{tabular}{l|ccc|cc|cc} 
    \toprule 
    \multicolumn{1}{c}{\multirow{3}{*}{\centering \textbf{Method}}} &  \multicolumn{3}{c}{\textbf{MNER}} & \multicolumn{2}{c}{\textbf{MRE}} & \multicolumn{2}{c}{\textbf{MEE}} \\
    \cmidrule(lr){2-4}\cmidrule(lr){5-6}\cmidrule(lr){7-8}
    &  \multicolumn{1}{c}{Twitter-15} & \multicolumn{1}{c}{Twitter-17}  & \multicolumn{1}{c}{SNAP}  & \multicolumn{1}{|c}{MNRE-V1} &\multicolumn{1}{c|}{MNRE-V2} & \multicolumn{1}{c}{M$^2$E$^2$ MED} &\multicolumn{1}{c}{M$^2$E$^2$ MEAE} \\
     \midrule
   UMT~\cite{yu2020improving} &   73.4 & 73.4 & - &  - & 65.2 & - & -\\
   UMGF~\cite{Zhang2021UMGF} &   74.9 & 85.5 & - &  - & - & - & -\\
   MEGA~\cite{zheng2021multimodal} & 72.4 & 84.4 & 66.4 & - & - & - & -\\
   RpBERT~\cite{sun2021rpbert} & 74.4  & - & 85.7  & - &  -  & - & - \\
   R-GCN~\cite{Zhao2022R-GCN} & 75.0 & 87.1 & - & - & - & - & -\\
   HVPNeT~\cite{Chen2022HVPNeT} &  75.3 &  86.8  & -  & 81.8 & - & - & -\\
   ITA~\cite{wang2022ita} & \underline{78.0} & 89.8 &90.2& - & 66.9 & - & -\\ 
   MoRe~\cite{Wang2022MoRE}   & 77.3 & 88.7 & 89.3 &- & 65.8 &- &-\\
   MNER-QG~\cite{Jia2023MNER-QG} & 75.0 & 87.3 & - & - & - & - & -\\
   HamLearning~\cite{Liu2023HamLearning} & 76.5 & 87.1 & - & - & - & - & -\\
   EviFusion~\cite{Liu2023Evidence} & 75.5 & 87.4 & - & - & - & - & -\\
   BGA-MNER~\cite{Chen2023BGA-MNER} & 76.3 & 87.7 & - & - & - & - & -\\
   WASE~\cite{li2020WASEandM2E2}   &- & - &-  &- &- & 50.8  & 19.2 \\
   Unicl~\cite{liu2022multimedia}  & - & - &-  &- &- &57.6 & 23.4\\
\midrule
UMIE-Base (Ours)  &  76.1  & 88.1 & 87.7 & 84.3 & 74.8 & 60.5 & 22.5\\ 
UMIE-Large (Ours) &  77.2 & \underline{90.7}  & \underline{90.5}  & \underline{85.0} & \underline{75.5} & \underline{61.0} & \underline{23.6}\\
UMIE-XLarge (Ours) &  \textbf{78.2} & \textbf{91.4}  & \textbf{91.0}  & \textbf{86.4} & \textbf{76.2} & \textbf{62.1} & \textbf{24.5}\\
       \bottomrule
    \end{tabular}

        \caption{Performance comparison on three multimedia information extraction tasks in F1 score (\%). 
    The best performance is marked on bold and the second best result is marked underline.}    \label{tab:main-result}
\end{table*}


\subsection{Main Results}
Table~\ref{tab:main-result} shows the comparison in detail.
Our UMIE achieves on-par or significantly better performances compared to the previous baselines..
In particular, the UMIE-XL achieves the SoTA performance on all six datasets, when compared to the best model in each dataset.
UMIE marginally outperforms previous SoTA methods on both MRE datasets, with more than a 10\% absolute F1 advantage on MNRE-V2.
This performance superiority strongly indicates the effectiveness and generality of UMIE in handling various multimodal information extraction tasks, also demonstrating the success of our proposed visual encoder, gated attention module, and training strategies.
More importantly, such a unified single checkpoint with capable performance across the board can serve as a reliable starting point for their various MIE tasks. Therefore, MIE practitioners do not need to train and maintain separate models for each specific scenario or dataset.

The M$^2$E$^2$ dataset has only a test set available while UMIE can generalize well on this M$^2$E$^2$, outperforming previous SoTA results by a large margin.
This demonstrates the strong generalization capability of UMIE and the effectiveness of multi-task training.
Further in-depth generalization evaluation on MNER and MRE tasks will be discussed in the following experiment.

In addition, the UMIE can achieve even better results by using more powerful backbone models.
For instance, transitioning from Base to Large, and then to XLarge size in the FLAN-T5 model family, consistently improves the performance.
This indicates the effectiveness and versatility of our model design and training strategies.


\begin{table}[t]
\centering
    \small
    \centering
    \begin{tabular}{l|c|c}
    \toprule
    \multicolumn{1}{c}{\multirow{1}{*}{\textbf{Method}}} &  \multicolumn{1}{c}{Twitter-17}  & \multicolumn{1}{c}{MNRE-V2} \\
    \midrule
   UMT~\cite{yu2020improving} & 60.9 & - \\
   UMGF~\cite{Zhang2021UMGF} & 60.9 & - \\
   FMIT~\cite{lu2022FMIT} & 64.4 & - \\
   CAT-MNER~\cite{Wang2022CAT-MNER} & 64.5 & - \\
   BGA-MNER~\cite{Chen2023BGA-MNER} & 64.9 & - \\
   \midrule
   ChatGPT & 57.5 & 35.2 \\
   GPT-4~\cite{openai2023gpt4} & 66.6 & 42.1 \\
   \midrule
   UMIE-Base (Ours)  &  66.8  & 67.3  \\
   UMIE-Large (Ours)  &  \underline{68.5}  & \underline{68.8}  \\ 
   UMIE-XLarge (Ours)  &  \textbf{69.9}  & \textbf{69.6}  \\
       \bottomrule
    \end{tabular}
        \caption{Performance comparison on zero-shot setting (\%). We report the results of each paper and adopt the results of ChatGPT and GPT-4 from~\citet{Chen2023CoT-MNER}.}    \label{tab:zero-shot}
\vspace{-1.0em}
\end{table}

\subsection{Zero-shot Generalization}

We evaluate the generalization ability of our UMIE models to unseen MIE tasks in the zero-shot setting. 
Specifically, we exclude the corresponding training set (e.g., Twitter-17 or MNRE-V2) and train a checkpoint with the rest of the training data.
The baselines are trained with the same rule of the training set exclusion.
Therefore, these models are evaluated purely on their ability to generalize from related tasks to unseen MNER (Twitter-17) and MRE (MNRE-V2) datasets.

Table~\ref{tab:zero-shot} reports final results and we could observe that our UMIE substantially outperforms other zero-shot baselines and such advantages continually enlarge as model size increases.
In particular, UMIE outperforms LLMs including ChatGPT and even GPT-4 in both MNER and MRE tasks, strongly indicating the effectiveness of our model.
Considering that the generalization is also verified in MEE, these experiments indicate the ability of UMIE to adapt and generalize to new tasks effectively, even without direct training on corresponding task data.
The high performance of UMIE in zero-shot setting can be attributed to instruction-following ability and multi-task learning on multiple MIE tasks which allow our model to transfer knowledge from one task to another.
This proves that our model is not only capable of learning from different sources of multimodal data but also successfully applying learned knowledge to unseen task data, underscoring its flexibility and versatility.

\begin{table}[t]
\centering

\begin{tabular}{cl} 
\toprule 
\textbf{MNER}& \multicolumn{1}{l}{\textbf{Task Instructor}} \\
 \midrule
I0& \makecell[l]{Given the entity types: person, location, \\miscellaneous, organization. 
Please extract\\ the specified entity type.}\\
\midrule
I1& \makecell[l]{Identity the following entity type from the \\given sentence: person, location,\\ miscellaneous,
organization.}\\
\midrule
I2& \makecell[l] {Please extract entity type in the sentence. \\ Option: person, location, miscellaneous,\\
organization.
}\\
\bottomrule
\end{tabular}
\caption{Three variants of MNER instructor.}
 \label{tab:task-instructor-mner-three}
 \vspace{-1.0em}
\end{table}

\subsection{Robustness to Instruction-Following}

An ideal model with instruction-following capability would understand and execute instructions correctly, regardless of their phrasing.
In such context, users can get decent results without carefully crafting the instructions, which requires time-consuming and tricky prompt engineering.
Therefore, to evaluate the robustness of the instruction-following ability of UMIE models, as shown in Table~\ref{tab:task-instructor-mner-three}, we provide UMIE models with three kinds of task instructors for the MNER task.
The results in Table~\ref{tab:robust_instruction} demonstrate that UMIE models have decent robustness to various instructions, maintaining consistently high performances with all model sizes.

\begin{table}[t]
\centering
    \small
    \centering
    \begin{tabular}{l|l|ccc}
    \toprule
      & \multicolumn{1}{c|}{\multirow{1}{*}{\textbf{Method}}} &  \multicolumn{1}{c}{Twitter-15}  & \multicolumn{1}{c}{Twitter-17} & {SNAP} \\
   \midrule
   \multirow{3}{*}{I0} & UMIE-Base   &  75.8  &                            88.0 & 87.8  \\
                      &   UMIE-Large   &  \underline{76.8} & \underline{90.7}  & \underline{90.1}\\
                      &   UMIE-XLarge   &  \textbf{78.0}  & \textbf{92.1}  & \textbf{91.7}  \\
      \midrule
    \multirow{3}{*}{I1} & UMIE-Base  &  75.9  & 88.1 & 87.6  \\
                       &  UMIE-Large   &  \underline{77.2}  & \underline{90.4}  & \underline{90.0}\\
                       &UMIE-XLarge   &  \textbf{77.9}  & \textbf{91.0}  & \textbf{90.8}  \\

    \midrule
    \multirow{3}{*}{I2} & UMIE-Base  &  76.0  & 88.2 & 87.5  \\
                       &  UMIE-Large   &  \underline{77.0}  & \underline{90.5}  & \underline{90.4}\\
                      &UMIE-XLarge   &  \textbf{78.2}  & \textbf{91.2}  & \textbf{91.0}  \\
       \bottomrule
    \end{tabular}
        \caption{ Performance of UMIE with three MNER instructors in F1 score (\%).}
    \label{tab:robust_instruction}
\end{table}



\begin{figure}[t]
    \small
    \centering
    \begin{subfigure}[b]{0.505\columnwidth}
       	
        \includegraphics[width=1.0\textwidth]{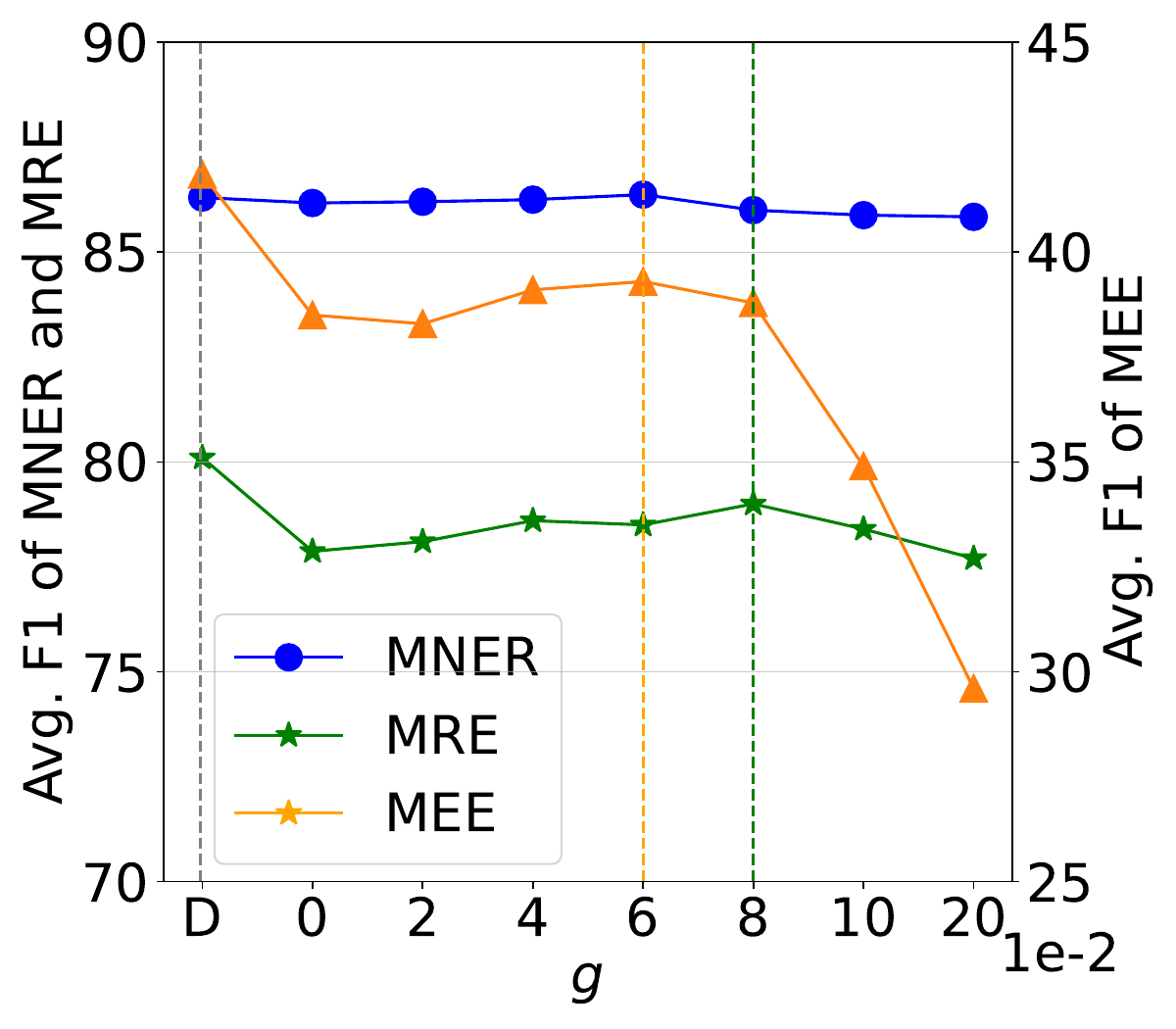}
         \subcaption{}
            \end{subfigure}
     \begin{subfigure}[b]{0.485\columnwidth}
    \includegraphics[width=1.0\textwidth]{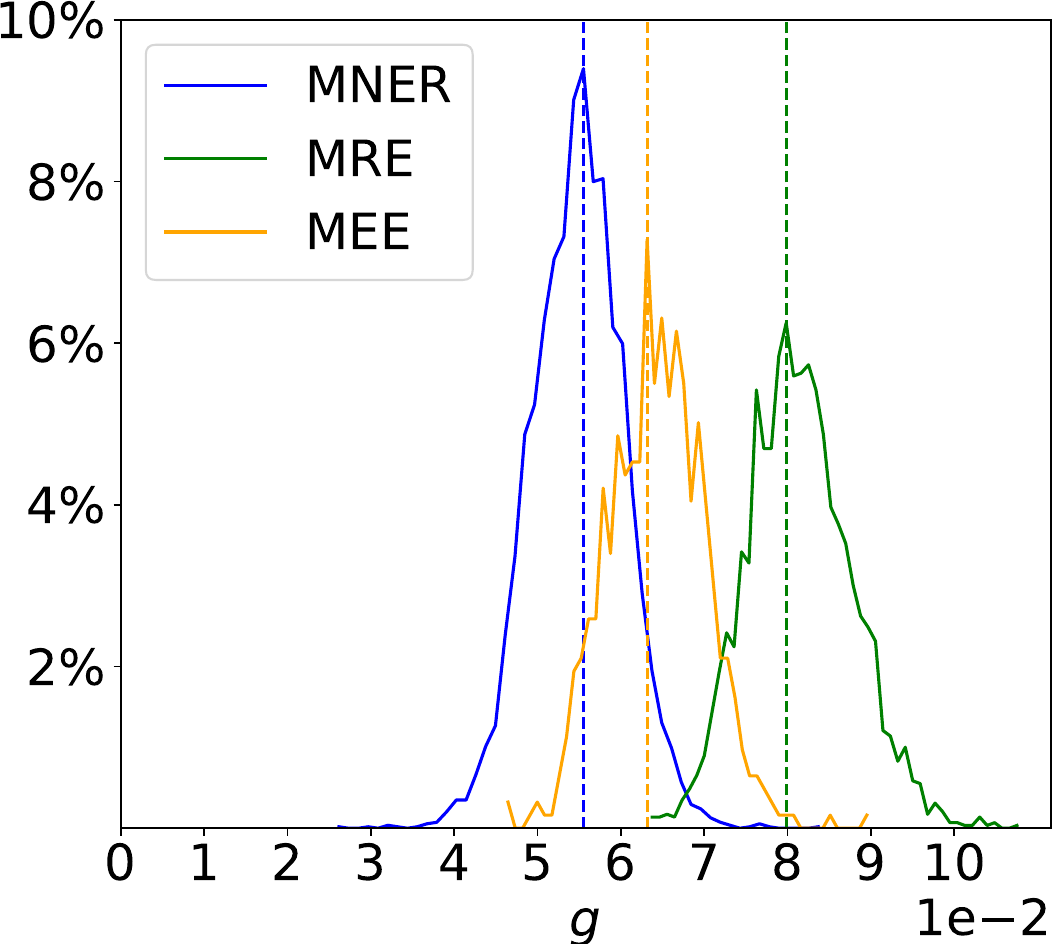}
          \subcaption{}
            \end{subfigure}
	\caption{(a) Performances of UMIE-Base averaged over three MIE tasks \textit{w.r.t} fixed gate value $g$ where D denotes the dynamic gate value of the UMIE;
 (b) $g$ value distribution of our dynamic gate module in three MIE tasks of UMIE-Base.
 }\label{fig:fix-attn-weight}
 \vspace{-1.5em}
\end{figure}

\subsection{Gate Control Ablation}
To disentangle the effects of the gated fusion mechanism and visual information in our UMIE model, we conduct an ablation study.
Concretely, we fix gate value $g$ in Eq.~\eqref{eqn: gate} for using visual information and observe the impact on our model's performance in Figure~\ref{fig:fix-attn-weight}.
Based on Figure~\ref{fig:fix-attn-weight}(b), it is evident that in the UMIE model, the effective range of the gate value is limited to 0-0.1. 
When the fixed gate value exceeds 0.1, a significant drop in performance is observed across all MIE tasks. Therefore, our primary focus is to present and analyze the performance specifically within the range of $g\in[0, 0.1]$.


In comparison to our dynamic gate value (D), any fixed gate value results in a decline in performance, particularly in MRE and MEE tasks. 
Specifically, when compared to $g=0$, representing only text without visual information, the dynamic gate value shows an average improvement of 0.1\% F1 score in MNER, 2.3\% in MRE, and 3.3\% in MEE.
This indicates that MRE and MEE are more heavily influenced by visual features, whereas MNER is hardly sensitive to visual aids.
This observation can be attributed to the fact that the MRE task involves inferring relationships between entities based on visual information, and the MEE task utilizes visual mentions as arguments, both of which require a high level of understanding of images.
Further evidence supporting this can be seen in the distribution of the $g$ values shown in Figure~\ref{fig:fix-attn-weight}(b). Specifically, the peak values of the histogram for MEE ($\approx$6.3e-2) and MRE ($\approx$8e-2) are higher compared to that of MNER ($\approx$5.5e-2), implying a stronger emphasis on visual features in the former two tasks.


Figure~\ref{fig:fix-attn-weight}(b) demonstrates the effectiveness of our gate module in assigning appropriate weights to each task. 
The peak values of the histogram for MRE ($\approx$6.3e-2) and MEE ($\approx$8e-2) coincide with the highest performances observed in the fixed gate value experiments, represented by the yellow dotted line for MEE and the green dotted line for MRE in Figure~\ref{fig:fix-attn-weight}(a).
By dynamically adjusting the weights for individual input instances based on their specific characteristics, our model achieves optimal results. This finding serves as strong evidence for the effectiveness and necessity of our dynamic gated fusion mechanism.

\subsection{Training Materials}
\begin{figure}[t]
    \small\centering\includegraphics[width=0.27\textwidth]{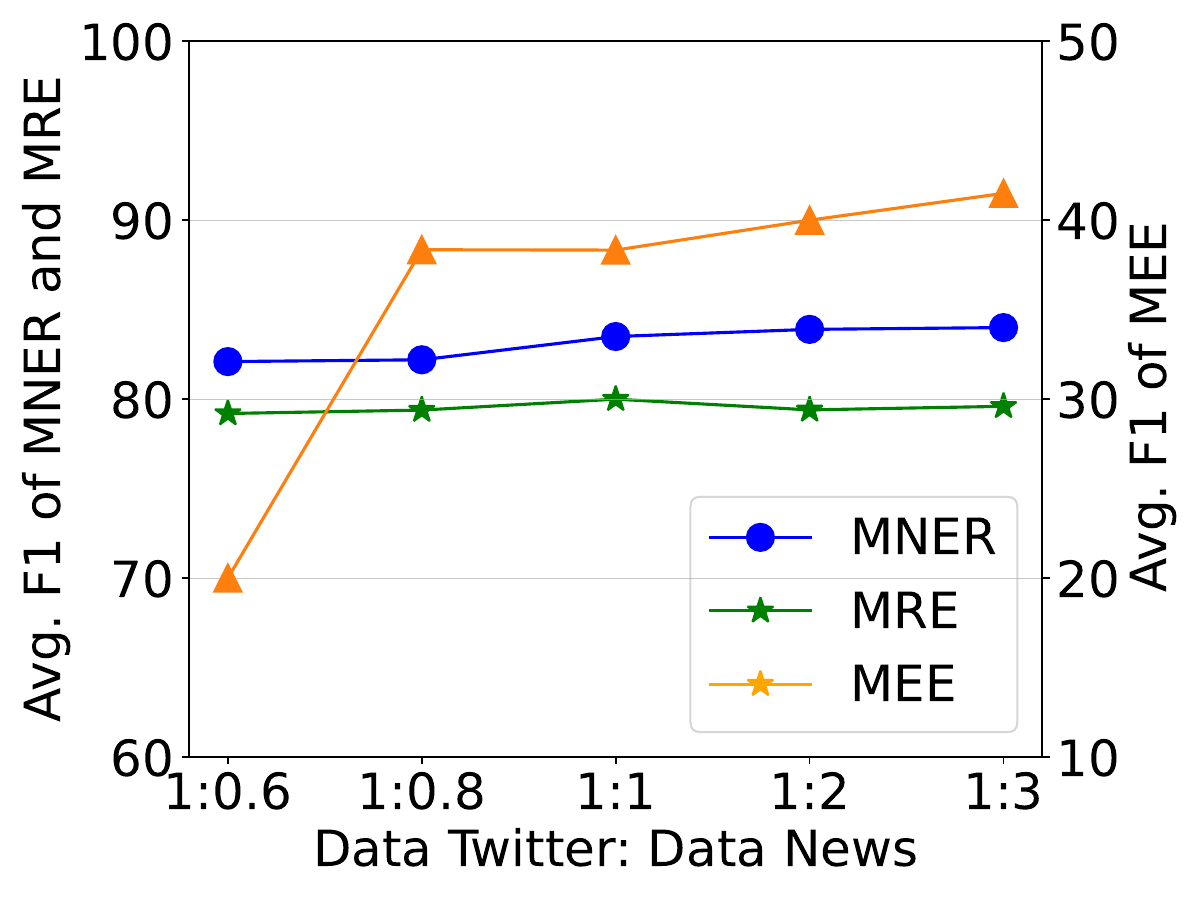}
    \caption{Averaged performance of three MIE tasks \textit{w.r.t} training sampling ratios of Twitter and News corpora.}\label{fig:fix-data-ratio}
\vspace{-1em}
\end{figure}

We also delve deep into the effects of training corpus, namely Twitter including Twitter-15, Twitter-17, SNAP, MNRE-V1, and MNRE-V2, and News including ACE2005 and M$^2$E$^2$, on the performance of the UMIE model. In Figure~\ref{fig:fix-data-ratio}, using UMIE-Base as an example, we observe that the Twitter corpus primarily contributes to MNER and MRE tasks, while the News corpus helps UMIE adapt to MEE.
Importantly, increasing training examples from the News domain does not negatively affect the tasks from the Twitter domain.
This indicates that UMIE series models exhibit strong compatibility when trained on cross-domain corpora and have the potential to benefit even more from larger-scale training materials.

\section{Conclusion}
This work proposes a unified framework for all MIE tasks with instruction tuning and multi-task learning based on encoder-decoder LLMs.
The gated attention module can effectively yield cross-modal information for decoding the MIE results.
Extensive experiments over three MIE tasks and six datasets show that our single UMIE model outperforms various prior task-specific SoTA methods across the board by a large margin, i.e., on average + 0.9\% for MNER, + 7.0\% for MRE, and + 2.8\% for MEE.
Our framework has strong generalization ability and robustness to instruction.
These desired properties make UMIE a foundation model in the MIE domain, demonstrating its broad applicability and promising potential for future work.

\section{Acknowledgements}
This work was supported  in part by Zhejiang Provincial Natural Science Foundation of China under Grant No. LGN22F020002 and Key Research and Development Program of Zhejiang Province under Grant No. 2022C03037.

\bibliography{custom}

\end{document}